\documentclass[final]{cvpr}

\usepackage{times}
\usepackage{epsfig}
\usepackage{graphicx}
\usepackage{amsmath}
\usepackage{amssymb}
\usepackage{booktabs}
\usepackage{multirow}
\usepackage{dcolumn}
\usepackage[utf8]{inputenc} 
\usepackage[T1]{fontenc}    
\usepackage{url}            
\usepackage{booktabs}       
\usepackage{amsfonts}       
\usepackage{nicefrac}       
\usepackage{microtype}      
\usepackage{lipsum}
\graphicspath{ {./images/} }
\usepackage{pbox}
\usepackage{epstopdf}
\usepackage{subfigure}
\usepackage{xspace}
\usepackage{comment}
\usepackage{bm}
\usepackage{bbm}
\usepackage{tabularx}
\usepackage{setspace}
\usepackage{sidecap}
\usepackage{xcolor}
\usepackage{wrapfig}
\usepackage{array}
\usepackage{xcolor}
\newcommand{\bx}{\textbf{x}}

\newcommand{\bg}{\textbf{g}}

\usepackage{times}
\usepackage{epsfig}
\usepackage{graphicx}
\usepackage{amsmath}
\usepackage{amssymb}
\usepackage{algorithm}
\usepackage{algorithmic}
\usepackage[utf8]{inputenc} 
\usepackage[T1]{fontenc}    
\usepackage{url}            
\usepackage{booktabs}       
\usepackage{amsfonts}       
\usepackage{nicefrac}       
\usepackage{microtype}      
\usepackage{pbox}
\usepackage{epstopdf}
\usepackage{subfigure}
\usepackage{xspace}
\usepackage{comment}
\usepackage{lipsum}
\usepackage{bm}
\usepackage{bbm}
\usepackage{tabularx}
\usepackage{setspace}
\usepackage{sidecap}
\usepackage{xcolor}
\usepackage{wrapfig}
\usepackage{multirow}
\usepackage{array}

\newcommand{\bt}{\textbf{t}}
\newcommand{\bv}{\textbf{v}}
\newcommand{\bbeta}{\bm{\beta}}
\newcommand{\bzeta}{\bm{\zeta}}
\newcommand{\balpha}{\bm{\alpha}}

\usepackage[pagebackref=true,breaklinks=true,letterpaper=true,colorlinks,bookmarks=false]{hyperref}

\usepackage[pagebackref=true,breaklinks=true,colorlinks,bookmarks=false]{hyperref}

\def\cvprPaperID{6784} 
\def\confYear{ICCV 2021}

\begin{document}

\title{Contrastive Multimodal Fusion with TupleInfoNCE}

\author{

\centerline{Yunze Liu$^{1,7}$ \quad Qingnan Fan$ ^{3}$ \quad Shanghang Zhang$ ^{4}$ \quad Hao Dong$ ^{5,6,8}$ \quad Thomas Funkhouser$ ^{2}$ \quad Li Yi$ ^{1,2}$\thanks{Corresponding author}}\\
\\
$^1$IIIS, Tsinghua University \quad $^2$Google Research \quad $^3$Stanford University \\
$^4$UC Berkeley \quad $^5$CFCS, CS Dept., Peking University \quad $^6$AIIT, Peking University\\
$^7$Xidian University \quad $^8$Peng Cheng Laboratory \\
{\tt\normalsize \{liuyzchina,fqnchina\}@gmail.com, \{ericyi,tfunkhouser\}@google.com}\\
{\tt\normalsize shz@eecs.berkeley.edu, hao.dong@pku.edu.cn}
}
\maketitle
\begin{abstract}
This paper proposes a method for representation learning of multimodal data using contrastive losses.  A traditional approach is to contrast different modalities to learn the information shared among them.  However, that approach could fail to learn the complementary synergies between modalities that might be useful for downstream tasks.  Another approach is to concatenate all the modalities into a tuple and then contrast positive and negative tuple correspondences.  However, that approach could consider only the stronger modalities while ignoring the weaker ones.  To address these issues, we propose a novel contrastive learning objective, TupleInfoNCE.  It contrasts tuples based not only on positive and negative correspondences, but also by composing new negative tuples using modalities describing different scenes. Training with these additional negatives encourages the learning model to examine the correspondences among modalities in the same tuple, ensuring that weak modalities are not ignored.  We provide a theoretical justification based on mutual-information for why this approach works, and we propose a sample optimization algorithm to generate positive and negative samples to maximize training efficacy.  We find that TupleInfoNCE significantly outperforms previous state of the arts on three different downstream tasks.
\end{abstract}
\section{Introduction}

\begin{figure}[t]
	\centering
	\includegraphics[width=\linewidth]{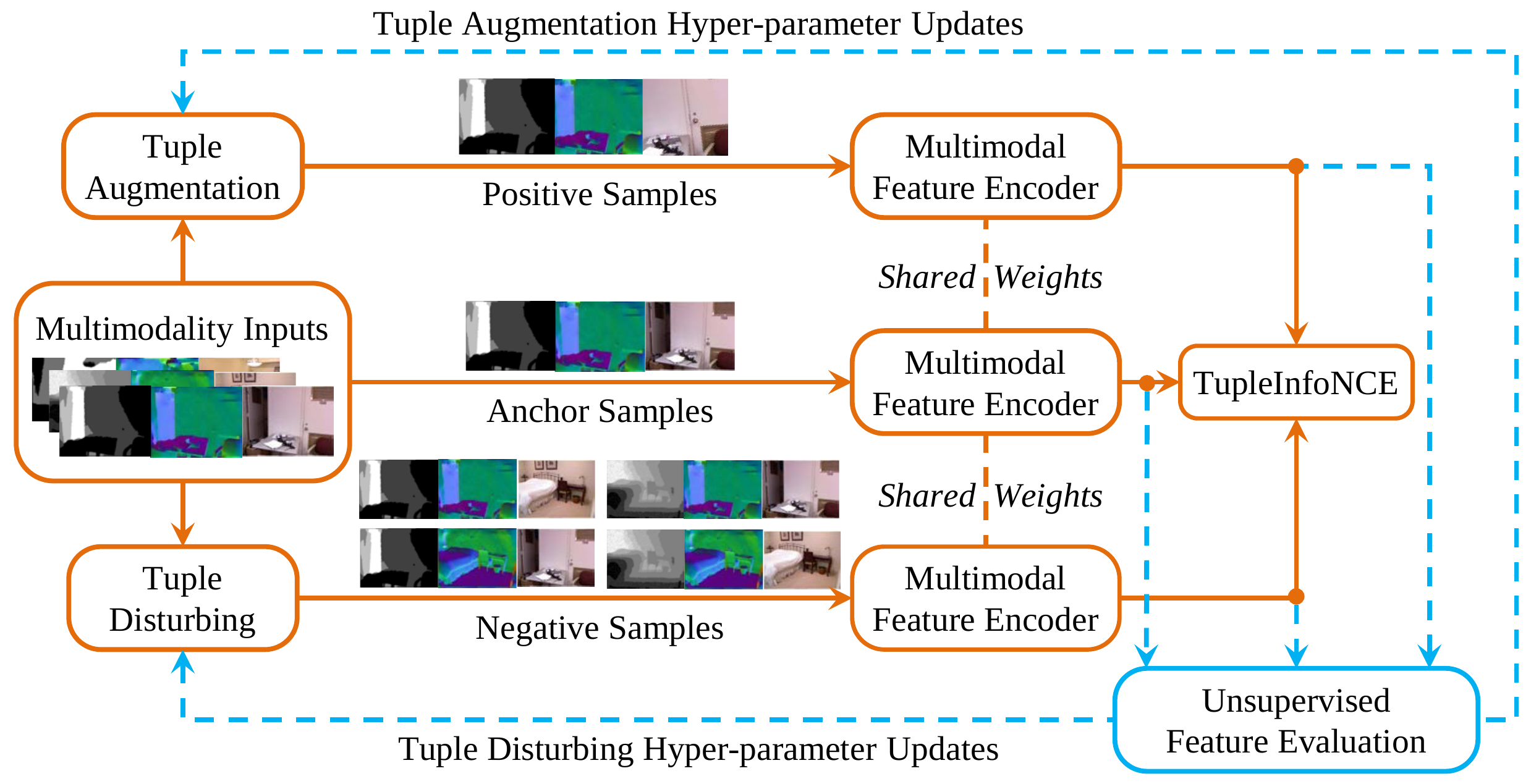}
	\caption{Overview of sample-optimized TupleInfoNCE.}
	\label{fig:teaser}
	\vspace{-15pt}
\end{figure}

Human perception of the world is naturally multimodal. What we see, hear, and feel all contain different kinds of information. Various modalities complement and disambiguate each other, forming a representation of the world. Our goal is to train machines to fuse such multimodal inputs to produce such representations in a self-supervised manner without manual annotations.

An increasingly popular self-supervised representation learning paradigm is contrastive learning, which learns feature representations via optimizing a contrastive loss and solving an instance discrimination task~\cite{oord2018representation, he2020momentum, chen2020simple}. Recently several works have explored contrastive learning for multimodal representation learning~\cite{tian2019contrastive, alayrac2020self, liu2020p4contrast}. Among them, the majority~\cite{tian2019contrastive, alayrac2020self} learn a crossmodal embedding space -- they contrast different modalities to capture the information shared across modalities. However, they do not examine the fused representation of multiple modalities directly, failing to fully leverage multimodal synergies. To cope with this issue,~\cite{liu2020p4contrast} proposes an RGB-D representation learning framework to directly contrast pairs of point-pixel pairs. However, it is restricted to two modalities only.

Instead of contrasting different data modalities, we propose to contrast multimodal input tuples, where each tuple element corresponds to one modality. We learn representations so that tuples describing the same scene (set of multimodal observations) are brought together while tuples from different scenes are pushed apart. This is more general than crossmodal contrastive learning. It not only supports extracting the shared information across modalities,
but also allows modalities to disambiguate each other and to keep their specific information, producing better-fused representations.

However, contrasting tuples is not as straightforward as contrasting single elements, especially if we want the learned representation to encode the information from each element in the tuple and to fully explore the synergies among them. The core challenge is: ``which tuple samples to contrast?'' Previously researchers~\cite{xie2020pointcontrast, liu2020p4contrast} have observed that always contrasting tuples containing corresponding elements from the same scene can converge to a lazy suboptimum where the network relies only on the strongest modality for scene discrimination. Therefore to avoid weak modalities being ignored and to facilitate modality fusion, we need to contrast from more challenging negative samples. Moreover, we need to optimize the positive samples as well so that the contrastive learning can keep the shared information between positive and anchor samples while abstracting away nuisance factors. Strong variations between the positive and anchor samples usually result in smaller shared information but a greater degree of invariance against nuisance variables. Thus a proper tradeoff is needed.

To handle the above challenges, we propose a novel contrastive learning objective named TupleInfoNCE (Figure~\ref{fig:teaser}). Unlike the popular InfoNCE loss~\cite{oord2018representation}, TupleInfoNCE is designed explicitly to facilitate multimodal fusion. TupleInfoNCE leverages positive samples generated via augmenting anchors and it exploits challenging negative samples whose elements are not necessarily in correspondence. These negative samples encourage a learning model to examine the correspondences among elements in an input tuple, ensuring that weak modalities and the modality synergy are not ignored. To generate such negative samples we present a tuple disturbing strategy with a theoretical basis for why it helps.

TupleInfoNCE also introduces optimizable hyper-parameters to control both the negative sample and the positive sample distributions. This allows optimizing samples through a hyper-parameter optimization process. We define reward functions regarding these hyper-parameters and measure the quality of learned representations via unsupervised feature evaluation. We put unsupervised feature evaluation in an optimization loop that updates these hyper-parameters to find a sample-optimized TupleInfoNCE (Figure \ref{fig:teaser}).

We evaluate TupleInfoNCE on a wide range of multimodal fusion tasks including multimodal semantic segmentation on NYUv2~\cite{silberman2012indoor}, multimodal object detection on SUN RGB-D~\cite{song2015sun} and multimodal sentiment analysis on CMU-MOSI~\cite{zadeh2016multimodal} and CMU-MOSEI~\cite{zadeh2018multimodal}. We demonstrate significant improvements over previous state-of-the-art multimodal self-supervised representation learning methods (+\textbf{4.7} mIoU on NYUv2, +\textbf{1.2} mAP@0.25 on SUN RGB-D, +\textbf{1.0\%} acc7 on MOSI, and +\textbf{0.5\%} acc7 on MOSEI).

Our key contributions are threefold. First, we present a novel TupleInfoNCE objective for contrastive multimodal fusion with a theoretical justification. Secondly, we pose the problem of optimizing TupleInfoNCE with a self-supervised approach to select the contrastive samples. Finally, we demonstrate state-of-the-art performance on a wide range of multimodal fusion benchmarks and provide ablations to evaluate the key design decisions.


\section{Related Work}
\subsection{Self-Supervised Multimodal Learning}
Self-supervised learning (SSL) uses auxiliary tasks to learn data representation from the raw data without using additional labels~\cite{vincent2008extracting,jakab2018unsupervised,misra2016shuffle,fernando2017self,wei2018learning}, helping to improve the performance of the downstream tasks.
Recently, research on SSL leverages multimodal properties of the data~\cite{chung2016out,arandjelovic2018objects,tian2019contrastive,hazarika2020misa,alayrac2020self,liu2020p4contrast}. The common strategy is to explore the natural correspondences among different views and use contrastive learning (CL) to learn representations by pushing views describing the same scene closer, while pushing views of different scenes apart~\cite{chung2016out,arandjelovic2018objects,tian2019contrastive,hazarika2020misa,alayrac2020self}. We refer to this line of methods as crossmodal embedding, which focuses on extracting the information shared across modalities rather than examining the fused representation directly, failing to fully explore the modality synergy for multimodal fusion. 

\subsection{Contrastive Representation Learning}

CL is a type of SSL that has received increasing attention for it brings tremendous improvements on representation learning. 
According to the learning method, it can be grouped into Instance-based~\cite{oord2018representation, he2020momentum, chen2020improved, chen2020simple} and Prototype-based CL~\cite{li2020prototypical, caron2020unsupervised}; 
According to the modality of data, it can be categorized into single-modality based~\cite{chuang2020debiased,ho2020contrastive} and multimodality based CL~\cite{alayrac2020self,liu2020p4contrast,tian2019contrastive}. 
An underexplored challenge for CL is how to select hard negative samples to build the negative pair~\cite{kalantidis2020hard,robinson2020contrastive,ho2020contrastive,chuang2020debiased}. Most existing methods either increase batch size or keep large memory banks, leading to large memory requirements~\cite{he2020momentum}.
Recently, several works study CL from the perspective of mutual information (MI).
\cite{tian2020makes} argues MI between views should be reduced by data augmentation while keeping task-relevant information intact. 
\cite{wu2020mutual} shows the family of CL algorithms maximizes a lower bound on MI between multi-``views” where typical views come from image augmentations, and finds the choice of negative samples and views are critical to these algorithms.  
We build upon this observation with an optimization framework for selecting contrastive samples.

\subsection{AutoML}
AutoML is proposed to automatically create models that outperform the manual design. The progress of neural architectural search (NAS)~\cite{zoph2018learning,liu2018progressive,baker2016designing}, data augmentation strategy search~\cite{cubuk2018autoaugment,lim2019fast} and loss function search~\cite{li2019lfs} have greatly improved the performance of neural networks.
But most of these methods focus on a supervised learning setting. Recently, developing AutoML techniques in an unsupervised/self-supervised learning scenario has drawn more attention~\cite{liu2020labels, tian2020makes,metzger2020evaluating}.
UnNAS~\cite{liu2020labels} shows the potential of searching for better neural architectures with self-supervision.
InfoMin~\cite{tian2020makes} and SelfAugment~\cite{metzger2020evaluating} explore how to search better data augmentation for CL on 2D images. In our work, we focus on optimizing two key components of a multimodal CL framework unsupervisedly - data augmentation and negative sampling strategies, none of which has been previously explored for generic multimodal inputs.

\section{Revisiting InfoNCE}
\label{sec:infonce}
Before describing our method, we first review the InfoNCE loss widely adopted for contrastive representation learning~\cite{oord2018representation}, and then discuss its limitations for multimodal inputs. Given an anchor random variable $\bx_{1,i}\sim p(\bx_1)$, the popular contrastive learning framework aims to differentiate a positive sample $\bx_{2,i}\sim p(\bx_2|\bx_{1,i})$ from negative samples $\bx_{2,j}\sim p(\bx_2)$. This is usually done by minimizing the InfoNCE loss:
\vspace{-2mm}
\begin{equation}
\mathcal{L_{\text{NCE}}}=-\mathbb{E}\left[\text{log}\frac{f(\bx_{2,i}, \bx_{1,i})}{\sum_{j=1}^Nf(\bx_{2,j}, \bx_{1,i})}\right]
\vspace{-1mm}
\end{equation}
where $f(\bx_{2,j},\bx_{1,i})$ is a positive scoring function usually chosen as a log-bilinear model. It has been shown that minimizing $\mathcal{L_{\text{NCE}}}$ is equivalent to maximizing a lower bound of the mutual information $I(\bx_2;\bx_1)$. Many negative samples are required to properly approximate the negative distribution $p(\bx_2)$ and tighten the lower bound.

In the problem setting of multimodal inputs, an input sample can be represented as a $K$-tuple $\bt=(\bv^1,\bv^2,...,\bv^K)$ where each element $\bv^k$ corresponds to one modality and $K$ denotes the total number of modalities being considered. A straightforward way of learning multimodal representations is to draw anchor samples $\bt_{1,i}\sim p(\bt_1)$, their positive samples $\bt_{2,i}\sim p(\bt_2|\bt_{1,i})$ and negative samples $\bt_{2,j}\sim p(\bt_2)$, and then optimize the InfoNCE objective. However, previous works~\cite{xie2020pointcontrast,liu2020p4contrast} observe that even when $K=2$ simply drawing negative samples from the marginal distribution $p(\bt_2)$ is insufficient for learning good representations. Weak modalities tend to be largely ignored and synergies among modalities are not fully exploited. The issue becomes more severe when $K>2$ when the informativeness of different modalities varies a lot. 

\begin{figure}[t]
	\centering
	\includegraphics[width=0.85\linewidth]{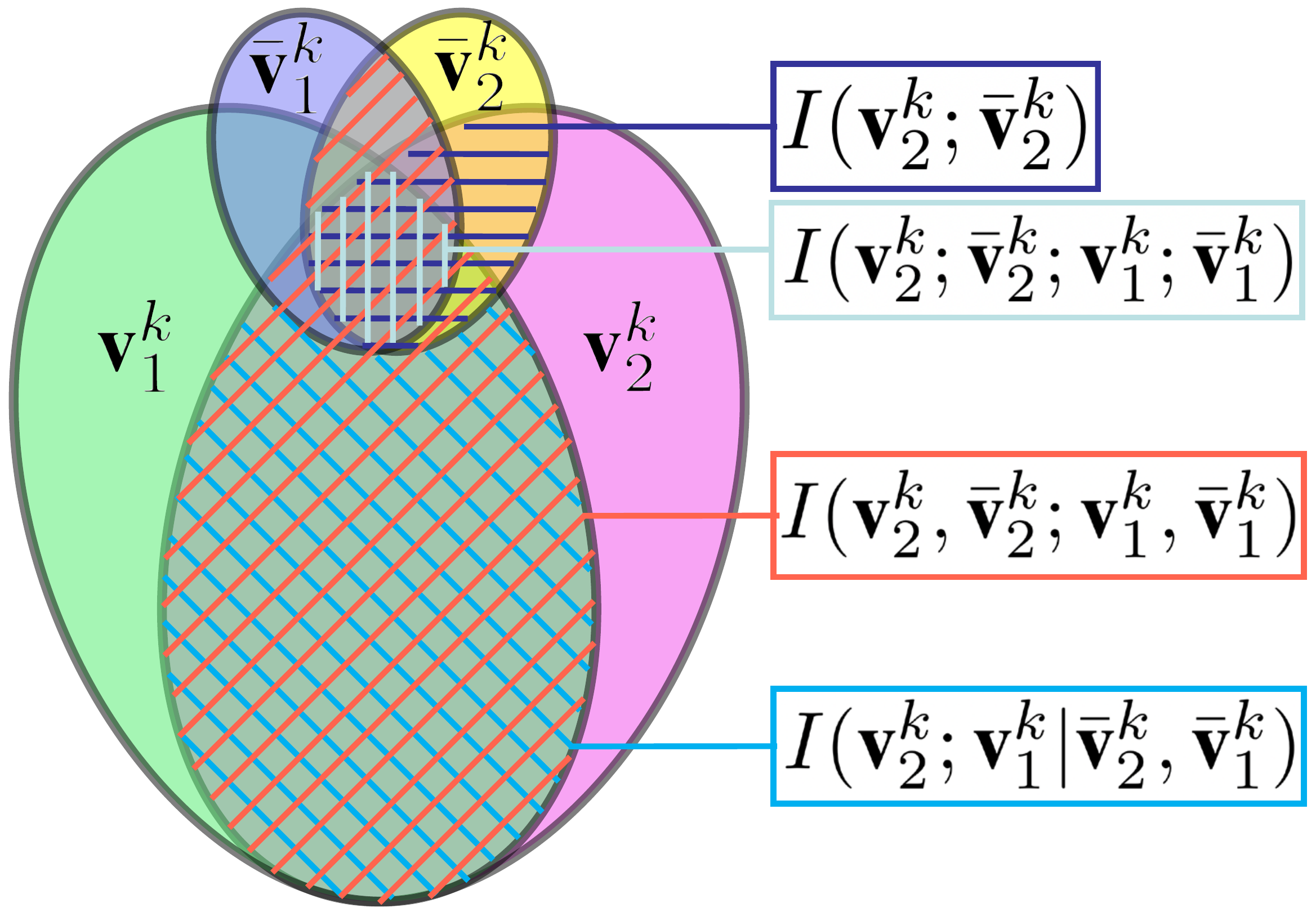}
	\caption{Information diagram}
	\label{fig:info_diag}
	\vspace{-5mm}
\end{figure}

Figure~\ref{fig:info_diag} provides an intuitive explanation. When one modality $\bv^k$ is particularly informative compared with the rest modalities $\bar{\bv}^k$ in the input tuple $\bt$, namely $I(\bv_2^k;\bv_1^k)\gg I(\bar{\bv}_2^k;\bar{\bv}_1^k)$, maximizing a lower bound of $I(\bt_2;\bt_1)=I(\bv_2^k, \bar{\bv}_2^k; \bv_1^k, \bar{\bv}_1^k)$ will be largely dominated by the modality specific information $I(\bv_2^k; \bv_1^k| \bar{\bv}_2^k, \bar{\bv}_1^k)$, which is usually not as important as the information shared across modalities $I(\bv_2^k; \bar{\bv}_2^k; \bv_1^k; \bar{\bv}_1^k)$. Overemphasizing the modality specific information from the strong modality might sacrifice the weak modalities and the modality synergy during learning.

\section{TupleInfoNCE}
\label{sec:tupleinfonce}
To alleviate the limitations of InfoNCE for overlooking weak modalities and the modality synergy, we present a novel TupleInfoNCE objective. We leverage a \textit{tuple disturbing} strategy to generate challenging negative samples, which prevents the network from being lazy and only focusing on strong modalities. In addition, we introduce optimizable data augmentations which are applied to anchor samples for positive sample generation. We optimize both the positive and negative samples to balance the information contributed by each modality. All these are incorporated into the proposed TupleInfoNCE objective, designed explicitly to facilitate multimodal fusion.

\subsection{Tuple disturbing and augmentation}

\noindent\textbf{Tuple disturbing} 
Generating challenging negative samples is fundamentally important to learning effective representation in contrastive learning, especially in the case of multimodal fusion setting where the strong modalities tend to dominate the learned representation \cite{liu2020p4contrast,xie2020pointcontrast}. 
We present a \textit{tuple disturbing} strategy to generate negative samples where not all modalities are in correspondence and certain modalities exhibit different scenes.

Given an anchor sample $(\bv_{1,i}^1,...,\bv_{1,i}^k,...,\bv_{1,i}^K)$ and its positive sample $(\bv_{2,i}^1,...,\bv_{2,i}^k,...,\bv_{2,i}^K)$, we propose a $k$-disturbed negative sample represented as $(\bv_{2,j}^1,...,\bv_{2,d(j)}^k,...,\bv_{2,j}^K)$, where $d(\cdot)$ is a disturbing function producing a random index from the sample set.  The negative sample has $K-1$ modalities $\bar{\bv}_{2,j}^k$ from one scene and one modality $\bv_{2,d(j)}^k$ from a different scene. 
Therefore, in order to correctly discriminate the positive sample from $k$-disturbed negative samples, the learned representation has to encode the information of the $k$-th modality, since the $K$-tuple could become negative only due to differences in the $k$-th modality.
$k$-disturbed negative samples become especially challenging when they are only partially negative, e.g. $\bar{\bv}_{2,j}^k$ becomes very similar to $\bar{\bv}_{2,i}^k$. Simply treating $\bv^k$ as an independent modality without considering its correlation with the rest modalities is not able to fully suppress the score of such partially negative samples in a log-bilinear model. Only when the network tells the disturbed modality $\bv_{2,d(j)}^k$ is not in correspondence with the rest modalities $\bar{\bv}_{2,j}^k$, can it fully suppress the partially negative samples. Therefore $k$-disturbed negative samples encourage the correlation between each modality and the rest to be explored.

We disturb each modality separately and generate $K$ types of negative samples to augment the vanilla InfoNCE objective. This enforces the representation learning of each specific modality in the multimodal inputs. We use $\alpha_k$ to represent the ratio of $k$-disturbed negative samples. Intuitively, the larger $\alpha_k$ we use, the more emphasis we put on the $k$-th modality.



\vspace*{2mm}\noindent\textbf{Tuple augmentation} Given an anchor sample $\bt_1$, we apply the data augmentation to each modality separately to generate the positive sample $\bt_2$.
The data augmentation applied to modality $\bv^k$ will directly influence $I(\bv_2^k;\bv_1^k)$~\cite{tian2020makes}, which roughly measures the information contribution of modality $\bv^k$ in $I(\bt_2;\bt_1)$. To further balance the contribution of each modality in our fused representation, we parameterize these data augmentations with a hyper-parameter $\bbeta$ and make $\bbeta$ optimizable for different modalities.

\subsection{Objective function}
\begin{figure}[t]
	\centering
	\includegraphics[width=0.9\linewidth]{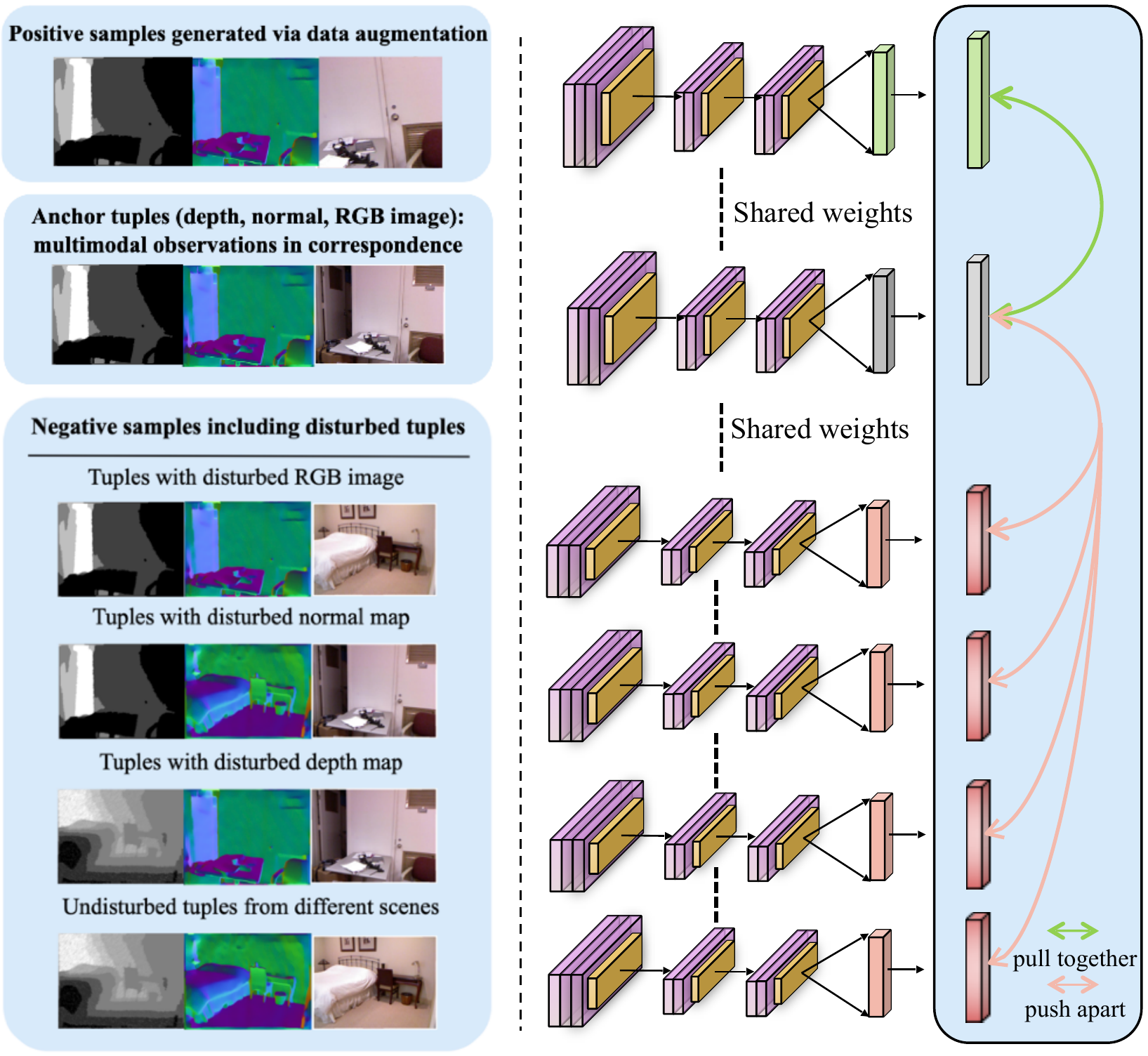}
	\caption{An example of the TupleInfoNCE objective for RGB, depth and normal map fusion.}
	\label{fig:tupleinfonce}
	\vspace{-6mm}
\end{figure}

The TupleInfoNCE objective is designed for fusing the multimodal input tuple $\bt=(\bv^1, \bv^2, ..., \bv^K)$. Given an anchor sample $\bt_{1,i}\sim p(\bt_1)$, we draw its positive sample $\bt_{2,i}\sim p_{\bbeta}(\bt_2|\bt_{1,i})$, and negative sample $\bt_{2,j|j\neq i}\sim q_{\balpha}(\bt_2)$ following a ``proposal'' distribution where either all modalities are in correspondence yet stem from a different scene, or each modality is disturbed to encourage modality synergy. To be specific, with probability $\alpha_0$ we sample negative samples from $p(\bt_2)$, and with probability $\alpha_k$ we sample $k$-disturbed negative samples from $p(\bar{\bv}_2^k)p(\bv_2^k)$, where $\{\alpha_k\}_{k=0}^K$ is a set of prior probabilities balancing different types of negative samples which sum to 1. This essentially changes our negative sample distribution to be $q_{\balpha}(\bt_2)=\alpha_0p(\bt_2)+\sum_{k=1}^K\alpha_kp(\bar{\bv}_2^k)p(\bv_2^k)$. Therefore, the TupleInfoNCE objective is defined as below:
\vspace{-2mm}
\begin{equation}
\mathcal{L}_{\text{TNCE}}^{\balpha\bbeta}=-\underset{\substack{\bt_{2,i}\sim p_{\bbeta}(\bt_2|\bt_{1,i})\\\bt_{2,j|j\neq i}\sim q_{\balpha}(\bt_2)}}{\mathbb{E}}\left[\text{log}\frac{f(\bt_{2,i}, \bt_{1,i})}{\sum_jf(\bt_{2,j}, \bt_{1,i})}\right]
\vspace{-2mm}
\end{equation}
where $f(\bt_{2,j}, \bt_{1,i})=\text{exp}(\bg(\bt_{2,j})\cdot \bg(\bt_{1,i})/\tau)$
and $\bg(\cdot)$ represents a multimodal feature encoder and $\tau$ is a temperature parameter. We provide an example for the TupleInfoNCE objective in Figure~\ref{fig:tupleinfonce}. The hyper-parameters $\balpha$ and $\bbeta$ can be optimized to allow flexible control over the contribution of different modalities as introduced in the next section.

\vspace*{2mm}\noindent\textbf{Connection with Mutual Information estimation} To better understand why $\mathcal{L}_{\text{TNCE}}^{\balpha\bbeta}$ is more suited for multimodal fusion than $\mathcal{L}_{\text{NCE}}$, we provide a theoretical analysis from the information theory perspective. As we mentioned in Section~\ref{sec:infonce}, minimizing $\mathcal{L}_{\text{NCE}}$ is equivalent to maximizing a lower bound of $I(\bt_2;\bt_1)$, which could lead to weak modalities and the modality synergy being ignored. Minimizing $\mathcal{L}_{\text{TNCE}}^{\balpha\bbeta}$, instead, is equivalent to maximizing a lower bound of $I(\bt_2;\bt_1|\bbeta)+\sum_{k=1}^K\alpha_kI(\bv_2^k;\bar{\bv}_2^k)$ (please see supplementary material for a proof). As is shown in Figure~\ref{fig:info_diag}, $I(\bv_2^k;\bar{\bv}_2^k)$ puts more emphasis on the information shared across modalities to encourage modality synergy and to avoid weak modalities being ignored. The ratio of $k$-disturbed negative samples $\alpha_k$ plays the role of balancing $I(\bv_2^k;\bar{\bv}_2^k)$ and $I(\bt_2;\bt_1|\bbeta)$. And the data augmentation parameters $\bbeta$ directly influence $I(\bt_2;\bt_1|\bbeta)$ and further balance the information contribution of each modality.

\subsection{Sample Optimization}
The hyper-parameters $\balpha$ and $\bbeta$ designed for tuple disturbing and augmentation play a key role in the TupleInfoNCE objective design. Each set of $\balpha$ and $\bbeta$ will correspond to one specific objective and fully optimizing $\mathcal{L}_{\text{TNCE}}^{\balpha\bbeta}$ will result in a multimodal feature encoder $\bg^{\balpha\bbeta}$. Manually setting these hyper-parameters is not reliable, motivating us to explore ways to optimize these hyper-parameters. There are mainly two challenges to be addressed. The first is the evaluation challenge: we need a way to evaluate the quality of the multimodal feature encoder $\bg^{\balpha\bbeta}$ in an unsupervised manner since most existing works have demonstrated that InfoNCE loss itself is not a good evaluator~\cite{tian2020makes,metzger2020evaluating}. The second is the optimization challenge: we need an efficient optimization strategy to avoid exhaustively examining different hyper-parameters and training the whole network from scratch repeatedly. 
We will explain how we handle these challenges to optimize the ratio $\balpha$ of different types of negative samples in Section~\ref{sec:negopt}, and the hyper-parameter $\bbeta$ of augmented positive samples in Section~\ref{sec:posopt}.
\vspace{-3mm}

\subsubsection{Optimizing negative samples}
\label{sec:negopt}

To evaluate the modality fusion quality in the learned representations unsupervisedly, we propose to use crossmodal discrimination as a surrogate task. To efficiently optimize $\balpha$, we adopt a bilevel optimization scheme alternating between optimizing $\balpha$ and optimizing the main $\mathcal{L}_{\text{TNCE}}^{\balpha\bbeta}$ objective with a fixed $\balpha$. We elaborate on these designs below.

\vspace*{2mm}\noindent\textbf{Crossmodal discrimination} TupleInfoNCE differs from the naive InfoNCE in that it emphasizes more on each modality $\bv^k$ as well as its mutual information $I(\bv^k;\bar{\bv}^k)$ with the rest modalities $\bar{\bv}^k$. In order to learn a good representation that properly covers $I(\bv^k;\bar{\bv}^k)$, we propose a novel surrogate task, \textit{crossmodal discrimination}, which looks for the corresponding $\bar{\bv}^k$ only by examining $\bv^k$ in a holdout validation set.
Mathematically, we first generate a validation set $\{\bt_m\}_{m=1}^M$ by drawing $M$ random tuples $\bt_m=(\bv_m^1, \bv_m^2,...,\bv_m^K)\sim p(\bt)$. For each modality $\bv_m^k$, its augmented version is represented as $\bv_m^{\prime k}\sim p_{\bzeta_k}(\bv^{\prime k}|\bv_m^k)$ following a data augmentation strategy parameterized by $\bzeta_k$. Then the crossmodal discrimination task is defined as, given any $\bv_n^{\prime k}$ sampled from the augmented validation set $\{\bv_m^{\prime k}\}_{m=1}^M$, finding its corresponding rest modalities $\bar{\bv}_n^k$ in the set $\{\bar{\bv}_m^k\}_{m=1}^M$. 
To solve this surrogate task, for any $\bv_n^{\prime k}$ sampled from the augmented validation set $\{\bv_m^{\prime k}\}_{m=1}^M$, we first compute its probability that corresponds to $\bar{\bv}_l^k$ as,
\vspace{-1mm}
\begin{equation}
\label{eq:pnl}
p_{nl}^k(\bg^{\balpha\bbeta})=\frac{\text{exp}(\bg^{\balpha\bbeta}(\bv_n^{\prime k})\cdot\bg^{\balpha\bbeta}(\bar{\bv}_l^k)/\tau)}{\sum_{m=1}^M\text{exp}(\bg^{\balpha\bbeta}(\bv_n^{\prime k})\cdot\bg^{\balpha\bbeta}(\bar{\bv}_m^k)/\tau)}
\vspace{-1mm}
\end{equation}
where $\bg^{\balpha\bbeta}(\cdot)$ represents our optimal multimodal feature encoder trained via optimizing $\mathcal{L}_{\text{TNCE}}^{\balpha\bbeta}$ and $\tau$ is a temperature parameter. 
Then the crossmodal discrimination accuracy for the $k$-th modality can be computed as
\vspace{-1mm}
\begin{equation}
\mathcal{A}^k(\bg^{\balpha\bbeta})=\sum_{n=1}^M\mathbbm{1}(n=\mathop{\arg}\mathop{\max}_l\,p_{nl}^k(\bg^{\balpha\bbeta}))/M
\vspace{-1mm}
\end{equation}
where $\mathbbm{1}(\cdot)$ is an indicator function. $\mathcal{A}^k(\bg^{\balpha\bbeta})$ roughly measures how much $I(\bv^k;\bar{\bv}^k)$ the encoder $\bg^{\balpha\bbeta}$ has captured and provides cues regarding how we should adjust $\alpha_k$ in the negative samples. We can then leverage the crossmodal discrimination accuracy to optimize $\balpha$ through maximizing the following reward:
\vspace{-2mm}
\begin{equation}
\label{eq:negreward}
\mathcal{R}(\balpha)=\sum_{k=1}^K\mathcal{A}^k(\bg^{\balpha\bbeta})
\vspace{-2mm}
\end{equation}
which properly balances the contribution of different modalities and has a high correlation with downstream semantic inference tasks as shown in Section~\ref{sec:analysis}. Notice to handle missing modalities in the crossmodal discrimination task, we adopt a \textit{dropout training} strategy as introduced in the supplemental material.


\vspace*{2mm}\noindent\textbf{Bilevel optimization}
Now we describe how to efficiently optimize $\mathcal{R}(\balpha)$ with one-pass network training. We write our optimization problem as below:
\vspace{-2mm}
\begin{equation}
\begin{split}
\text{maximize}\,\mathcal{R}(\balpha)=\sum_{k=1}^K\mathcal{A}^k(\bg^{\balpha\bbeta})\\
\text{s.t.}\quad \bg^{\balpha\bbeta}=\mathop{\arg}\underset{\bg}{\mathop{\min}}\,\mathcal{L}_{\text{TNCE}}^{\balpha\bbeta}(\bg)
\end{split}
\vspace{-3mm}
\end{equation}

This is a standard bilevel optimization problem. Inspired by~\cite{li2019lfs}, we adopt a hyper-parameter optimization strategy which alternatively optimizes $\balpha$ and $\bg$ in a single training pass. Specifically, we relax the constraint that $\sum_{k=0}^K\alpha_k=1$ during the optimization and use an independent multivariate Gaussian $\mathcal{N}(\bm{\mu}_0,\sigma I)$ to initialize the distribution of $\balpha$. At each training epoch $t$, we sample B hyper-parameters $\{\balpha_1,...\balpha_B\}$ from distribution $\mathcal{N}(\bm{\mu}_t,\sigma I)$ and train our current feature encoder $\bg_t$ separately to generate B new encoders $\{\bg_{t+1}^1,...,\bg_{t+1}^B\}$. We evaluate the reward for each of these encoders on the validation set and update the distribution of $\balpha$ using REINFORCE~\cite{williams1992simple} as below:
\begin{equation}
\label{eq:reinforce}
\bm{\mu}_{t+1}=\bm{\mu}_t+\eta\frac{1}{B}\sum_{i=1}^BR(\balpha_i)\nabla_{\balpha}\text{log}(p(\balpha_i;\bm{\mu},\sigma))
\end{equation}
where $p(\balpha_i;\bm{\mu},\sigma)$ represents the PDF of the Gaussian distribution. We then pick up the encoder with the highest reward as our $\bg_{t+1}$ and continue with the next epoch. We repeat the above process until convergence.

\begin{algorithm}[tb]
	\caption{:Sample Optimization}
	\label{alg:SOTNCE}
	\begin{algorithmic}
		\STATE {\bfseries Input:} Initialized multimodal feature encoder $\bg_0$, initialized distribution $(\bm{\mu}_0^{\balpha},\sigma^{\balpha})$ and $(\bm{\mu}_0^{\bbeta},\sigma^{\bbeta})$, total training epochs $T$, distribution learning rate $\eta$
		\STATE {\bfseries Output:} Final multimodal feature encoder $\bg_T^{\balpha^*\bbeta^*}$
		\FOR{$t=1$ {\bfseries to} $T$}
		\IF{$t$ is even}
		\STATE Sample $B$ {\bfseries sampling ratio} hyper-parameters $\{\balpha_i\}_{i=1}^B$ via distribution $\mathcal{N}(\boldsymbol{\mu}_t^{\balpha},\sigma^{\balpha} I)$;
		\STATE  Train $\bg_t$ for one epoch separately with each $\balpha_i$ and get $\{\bg_{t+1}^i\}_{i=1}^B$;
		\STATE  Calculate rewards $\{\mathcal{R}(\balpha_i)\}_{i=1}^B$ using Equation~\ref{eq:negreward};
		\STATE  Decide the best model $i = \mathop{\arg}\mathop{\max}_{j} \mathcal{R}(\balpha_j)$;
		\STATE  Update $\bm{\mu}_{t+1}^{\balpha}$ using Equation~\ref{eq:reinforce};
		\STATE  Update $\bg_{t+1} = \bg_{t+1}^i$;
		\ELSIF{$t$ is odd}
		\STATE Sample $B$ {\bfseries data augmentation} hyper-parameters $\{\bbeta_i\}_{i=1}^B$ via distribution $\mathcal{N}(\boldsymbol{\mu}_t^{\bbeta},\sigma^{\bbeta} I)$;
		\STATE  Train $\bg_t$ for one epoch separately with each $\bbeta_i$ and get $\{\bg_{t+1}^i\}_{i=1}^B$
		\STATE  Calculate rewards $\{\mathcal{R}(\bbeta_i)=\}_{i=1}^B$ using Equation~\ref{eq:posreward};
		\STATE  Decide the best model $i = \mathop{\arg}\mathop{\max}_{j} \mathcal{R}(\bbeta_j)$;
		\STATE  Update $\bm{\mu}_{t+1}^{\bbeta}$ using Equation~\ref{eq:reinforce};
		\STATE  Update $\bg_{t+1} = \bg_{t+1}^i$;
		\ENDIF
		\ENDFOR
		\STATE return $\bg_T$
	\end{algorithmic}
\end{algorithm}

\subsubsection{Optimizing positive samples}
\label{sec:posopt}

Similar to optimizing $\balpha$, a reward function is required to evaluate our feature encoder $\bg^{\balpha\bbeta}$ in an unsupervised manner with respect to $\bbeta$. A straightforward approach is to adopt the total crossmodal discrimination accuracy defined in Equation~\ref{eq:negreward}. Through experiments, we observe two phenomena making this simple adaptation fail to optimize $\bbeta$ effectively.
We use $\bbeta$ and $\bzeta$ to represent the data augmentation parameters for training and validation respectively, and they do not have to be the same. 1). If we manually set $\bzeta$ to be fixed, the optimal $\bbeta$ maximizing the total accuracy highly correlates with $\bzeta$ and fails to generate truly good positive samples. 2). If we set $\bzeta$ to be the same as $\bbeta$ and optimize them together, we usually achieve the best total accuracy when no data augmentation is applied, though it has been shown a certain level of data augmentation is important for contrastive learning~\cite{chen2020simple,tian2020makes}. Therefore a better reward function is required for $\bbeta$ optimization.

We re-write our total crossmodal discrimination accuracy as $\sum_{k=1}^K\mathcal{A}^k(\bg^{\balpha\bbeta},\bzeta)$ to reflect the influence from $\bzeta$. Instead of manually setting $\bzeta$ which produces a chicken-and-egg problem for hyper-parameter optimization, we set $\bzeta=\bbeta$ and only optimize $\bbeta$. We follow the conclusion in~\cite{tian2020makes} and aim to use strong data augmentations, which reduces the information contribution by each modality but make the contributed information more robust to nuanced input noises. 
We observe that the total accuracy will decrease as we use stronger augmentations, and minimizing $\sum_{k=1}^K\mathcal{A}^k(\bg^{\balpha\bbeta},\bbeta)$ with respect to $\bbeta$ will effectively increase the augmentation magnitude. 
However, as discussed in~\cite{tian2020makes}, we should not increase the data augmentation without any constraints and there is a sweet spot going beyond which a larger data augmentation could harm the representation learning.
We find $\left\lVert\bbeta-\bzeta^*(\bbeta)\right\rVert^2$ providing cues for identifying the sweet spot, where $\bzeta^*(\bbeta)=\mathop{\arg}\mathop{\max}_{\bzeta}\sum_{k=1}^K\mathcal{A}^k(\bg^{\balpha\bbeta},\bzeta)$ represents the best $\bzeta$ maximizing the total crossmodal discrimination accuracy $\sum_{k=1}^K\mathcal{A}^k$ for a feature encoder trained with $\bbeta$. When $\bbeta$ is weak, we empirically discover that $\bzeta^*(\bbeta)$ is very close to $\bbeta$; when $\bbeta$ is too strong, smaller augmentation parameters on the validation set will lead to higher total accuracy, therefore leading to a large difference between $\bbeta$ and $\bzeta^*(\bbeta)$. We provide empirical studies supporting these findings in Section~\ref{sec:analysis}. Motivated by the above observations, we design our reward function as:
\vspace{-2mm}
\begin{equation}
\label{eq:posreward}
\begin{split}
\mathcal{R}(\bbeta)=1-\sum_{k=1}^K\frac{\mathcal{A}^k(\bg^{\balpha\bbeta},\bbeta)}{K}-\lambda\frac{\left\lVert\bbeta-\bzeta^*(\bbeta)\right\rVert^2}{\left\lVert\bbeta^{\text{max}}\right\rVert^2}
\end{split}
\vspace{-2mm}
\end{equation}
where $\lambda$ is a balancing parameter and $\bbeta^{\text{max}}$ denotes a predefined augmentation parameter upper bound used for the normalization purpose.

$\mathcal{R}(\bbeta)$ can be optimized in the same way as how $\mathcal{R}(\balpha)$ is optimized, and we alternate between optimizing $\bbeta$ and $\bg$ in a single training pass. We further combine the optimization of $\mathcal{R}(\balpha)$, $\mathcal{R}(\bbeta)$, and the multimodal encoder $\bg$ in Algorithm~\ref{alg:SOTNCE}, where we update $\balpha$ when the epoch number is even and update $\bbeta$ otherwise.

\section{Experiment}
In this section, we evaluate our method by transfer learning, i.e., fine-tuning on downstream tasks and datasets. Specifically, we first pretrain our backbone on each dataset without any additional data using the proposed TupleInfoNCE. Then we use the pre-trained weights as initialization and further refine them for target downstream tasks. In this case, good features could directly lead to performance gains in downstream tasks.

We present results for three popular multi-modality tasks: semantic segmentation on NYUv2~\cite{silberman2012indoor}, 3D object detection on SUN RGB-D~\cite{song2015sun}, and sentiment analysis on MOSEI~\cite{zadeh2018multimodal} and MOSI~\cite{zadeh2016multimodal} in Section~\ref{sec:NYU}, ~\ref{sec:SUNRGBD} and ~\ref{sec:MSA} respectively. In Section~\ref{sec:analysis}, extensive ablation studies, analysis and visualization are provided to justify design choices of our system.

\subsection{NYUv2 Semantic Segmentation}
\label{sec:NYU}
\textbf{Setup.} We first conduct experiments on NYUv2~\cite{silberman2012indoor} to see whether our method can help multimodal semantic scene understanding.
NYUv2 contains 1,449 indoor RGB-D images, of which 795 are used for training and 654 for testing. We use three modalities in this task: RGB, depth, and normal map. The data augmentation strategies we adopted include random cropping, rotation, and color jittering. We use ESANet~\cite{seichter2020efficient}, an efﬁcient ResNet-based encoder, as our backbone. We use the common 40-class label setting and mean IoU(mIoU) as the evaluation metric. 

We compare our method with the train-from-scratch baseline as well as the latest self-supervised multimodal representation learning methods including CMC~\cite{tian2019contrastive}, MMV FAC~\cite{alayrac2020self} and MISA~\cite{hazarika2020misa}, which are all based upon crossmodal embedding.
In addition, we include an InfoNCE~\cite{oord2018representation} baseline where we directly contrast multimodal input tuples without tuple disturbing and sample optimization. We also include supervised pretraining~\cite{seichter2020efficient} methods for completeness.

\begin{table}[h]
	\centering
	\caption{Semantic Segmentation results on NYUv2.}
	\label{tab:NYU}
	\newcolumntype{Y}{>{\centering\arraybackslash}X}
	{
		\setlength{\tabcolsep}{0.2em}
		\begin{tabularx}{\columnwidth}{>{\centering} m{0.65\columnwidth}|Y}
			\toprule
			Methods & mIoU \\
			\hline 
			Train from scratch &40.1  \\
			\hline
			Supervised pretrain on Imagenet  &50.3  \\
			\hline
			Supervised pretrain on Scenenet &51.6  \\
			\hline
			\hline
			CMC & 41.9 \\
			\hline
			MMV FAC &42.5\\
			\hline
			MISA &43.4 \\
			\hline
			InfoNCE &42.1 \\
			\hline
			Ours &\textbf{48.1} \\
			
			\bottomrule
		\end{tabularx}
	}
	\vspace{-10pt}
\end{table}

\textbf{Results.} Table~\ref{tab:NYU} shows that the previous best performing method MISA~\cite{hazarika2020misa} improves the segmentation mIoU by $3.3\%$ over the train-from-scratch baseline. When using InfoNCE~\cite{oord2018representation}, the improvement drops to $2.0\%$. Our method achieves $8.0\%$ improvement over the train-from-scratch baseline. The improvement from $40.1\%$ to $48.1\%$ confirms that we can produce better-fused representations to boost the segmentation performance on RGB-D scenes. Notably, our proposed TupleNCE, though only pretrained on NYUv2 self-supervisedly, is only \textasciitilde$3\%$ lower than supervised pretraining methods.

\subsection{SUN RGB-D 3D Object Detection}
\label{sec:SUNRGBD}

\textbf{Setup.} Our second experiment investigates how TupleInfoNCE can be used for 3D object detection in the SUN RGB-D dataset~\cite{song2015sun}. SUN RGB-D contains a training set with \textasciitilde5K single-view RGB-D scans and a test set with \textasciitilde5K scans. The scans are annotated with amodal 3D-oriented bounding boxes for objects from 37 categories. We use three modalities in this experiment: 3D point cloud, RGB color and height. Data augmentation used here is rotation for point cloud, jittering for RGB color, and random noise for height. We use VoteNet~\cite{qi2019deep} as our backbone, which leverages PointNet++~\cite{qi2017pointnet++} to process depth point cloud and supports appending RGB or height information as additional inputs. We compare our method with baseline methods including InfoNCE~\cite{oord2018representation}, CMC~\cite{tian2019contrastive}, and MISA~\cite{hazarika2020misa}. We use mAP@0.25 as our evaluation metric.

\begin{table}[h]
	\centering
	\caption{3D Object Detection results on SUN RGB-D.}
	\label{tab:SUNRGB-D}
	\newcolumntype{Y}{>{\centering\arraybackslash}X}
	{
		\setlength{\tabcolsep}{0.2em}
		\begin{tabularx}{\columnwidth}{>{\centering} m{0.70\columnwidth}|Y}
			\toprule
			Methods & mAP@0.25 \\
			\hline 
			Train from scratch & 56.3 \\
			\hline
			\hline 
			InfoNCE  & 56.8\\
			\hline
			CMC & 56.5 \\
			\hline
			MISA & 56.7\\
			\hline
			Ours  &\textbf{58.0} \\
			
			\bottomrule
		\end{tabularx}
	}
	\vspace{-10pt}
\end{table}

\textbf{Results.} Table~\ref{tab:SUNRGB-D} shows the object detection results. We find that previous self-supervised methods seem to struggle with 3D tasks: CMC and MISA achieve very limited improvement over the baseline trained from scratch. The improvement of InfoNCE~\cite{oord2018representation} is also very marginal (0.5\%), presumably because overemphasizing the modality-specific information from strong modalities might sacrifice the weak modalities as well as the modality synergy during learning. In contrast, TupleInfoNCE achieves $1.7\%$ $\text{mAP}$ improvement over the baseline trained from scratch, which more than triples the improvement InfoNCE achieved. The comparison between our method and InfoNCE directly validates the efficacy of the proposed TupleInfoNCE objective and sample optimization mechanism.

\subsection{Multimodal Sentiment Analysis}
\label{sec:MSA}
\textbf{Setup.} Our third experiment investigates multimodal sentiment analysis with the MOSI~\cite{zadeh2016multimodal} and MOSEI~\cite{zadeh2018multimodal} datasets, both providing word-aligned multimodal signals (language, visual and acoustic) for each utterance. MOSI contains 2198 subjective utterance-video segments. The utterances are manually annotated with a continuous opinion score between [-3,3], where -3/+3 represents strongly negative/positive sentiments. MOSEI is an improvement over MOSI with a higher number of utterances, greater variety in samples, speakers, and topics. Following the recent state-of-the-art multimodal self-supervised representation learning method MISA~\cite{hazarika2020misa}, we use features pre-extracted from the original raw data, which does not permit an intuitive way for data augmentation. Therefore we only optimize negative samples in this experiment. We use the same backbone as MISA~\cite{hazarika2020misa} to make a fair comparison. We use binary accuracy (Acc-2), 7-class accuracy (Acc-7), and F-Score as our evaluation metrics.

\textbf{Results.} As shown in Table~\ref{tab:MOSI} and~\ref{tab:MOSEI}, our method consistently outperforms previous methods on these very challenging and competitive datasets -- e.g., compared with the previous best performing method MISA, the Acc-7 goes up from 42.3 to 43.3 on MOSI, and from 52.2 to 52.7 on MOSEI. As these two approaches share the same network backbone and only differ in their strategy to learn the fused representation, the improvement provides strong evidence for the effectiveness of our method.

\begin{table}[h]
	\centering
	\caption{Multimodal sentiment analysis results on MOSI.}
	\label{tab:MOSI}
	\newcolumntype{Y}{>{\centering\arraybackslash}X}
	{
		\setlength{\tabcolsep}{0.2em}
		\begin{tabularx}{\columnwidth}{>{\centering} m{0.50\columnwidth}|Y|Y|Y}
			\toprule
			Methods & Acc-2 & Acc-7 & F-Score \\
			\hline 
			Train from scratch &83.0 &40.0&82.8 \\
			\hline
			\hline
			CMC & 83.3 & 39.5 &83.0 \\
			\hline
			MMV FAC &83.5& 41.5 &83.4\\
			\hline
			MISA &83.4 &42.3 &83.6 \\
			\hline
			InfoNCE &83.1  &40.5 &82.8\\
			\hline
			Ours &\textbf{83.6} &\textbf{43.3} &\textbf{83.8}\\
			\bottomrule
		\end{tabularx}
	}
	\vspace{-10pt}
\end{table}

\begin{table}[h]
	\centering
	\caption{Multimodal sentiment analysis results on MOSEI.}
	\label{tab:MOSEI}
	\newcolumntype{Y}{>{\centering\arraybackslash}X}
	{
		\setlength{\tabcolsep}{0.2em}
		\begin{tabularx}{\columnwidth}{>{\centering} m{0.50\columnwidth}|Y|Y|Y}
			\toprule
			Methods & Acc-2 & Acc-7 & F-Score \\
			\hline 
			Train from scratch &82.5&51.8&82.3 \\
			\hline
			\hline
			CMC & 83.3 & 50.8 &84.1 \\
			\hline
			MMV FAC &85.1& 52.0 &85.0\\
			\hline
			MISA &85.5 &52.2 &85.3 \\
			\hline
			InfoNCE &83.5  &52.0 &83.4\\
			\hline
			Ours &\textbf{86.1} &\textbf{52.7} &\textbf{86.0}\\
			\bottomrule
		\end{tabularx}
	}
	\vspace{-10pt}
\end{table}

\subsection{Further Analysis and Discussions}
\label{sec:analysis}

\noindent\textbf{Efficacy of sample optimization}
We run ablation studies with and without sample optimization to quantify its efficacy. We find that uniformly setting $\alpha_k$ without optimizing negative samples results in a 1.7\% mIoU drop on the NYUv2 semantic segmentation task, 0.5 mAP drop on the SUN RGB-D 3D object detection task, 0.6 Acc-7 drop on MOSI, and 0.4 Acc-7 drop on MOSEI. Manually designing data augmentation strategies without optimizing positive samples as in~\cite{tian2019contrastive} results in a 1.1 mIoU drop on NYUv2 and 0.6 mAP drop on SUN RGB-D. We also examine the optimized negative sampling strategy as well as the data augmentation strategy. On the NYUv2 dataset, we find the best performing negative sampling ratio among RGB, depth and normal is roughly $2:1:1$, showing that RGB is emphasized more in the fused representations. As for the data augmentation strategy, though we use the same types of data augmentations for all the three modalities on NYUv2, the optimal augmentation parameters vary from modality to modality. Considering image rotation with the hyper-parameter representing the rotation angle, we found that 40 degrees is the best hyper-parameter for RGB images, while 10 degrees is the best for depth and normal maps. Please refer to the supplementary material for more analysis regarding SUN RGB-D, MOSI, and MOSEI.

\begin{figure}[t]
	\centering
	\includegraphics[width=1.0\linewidth]{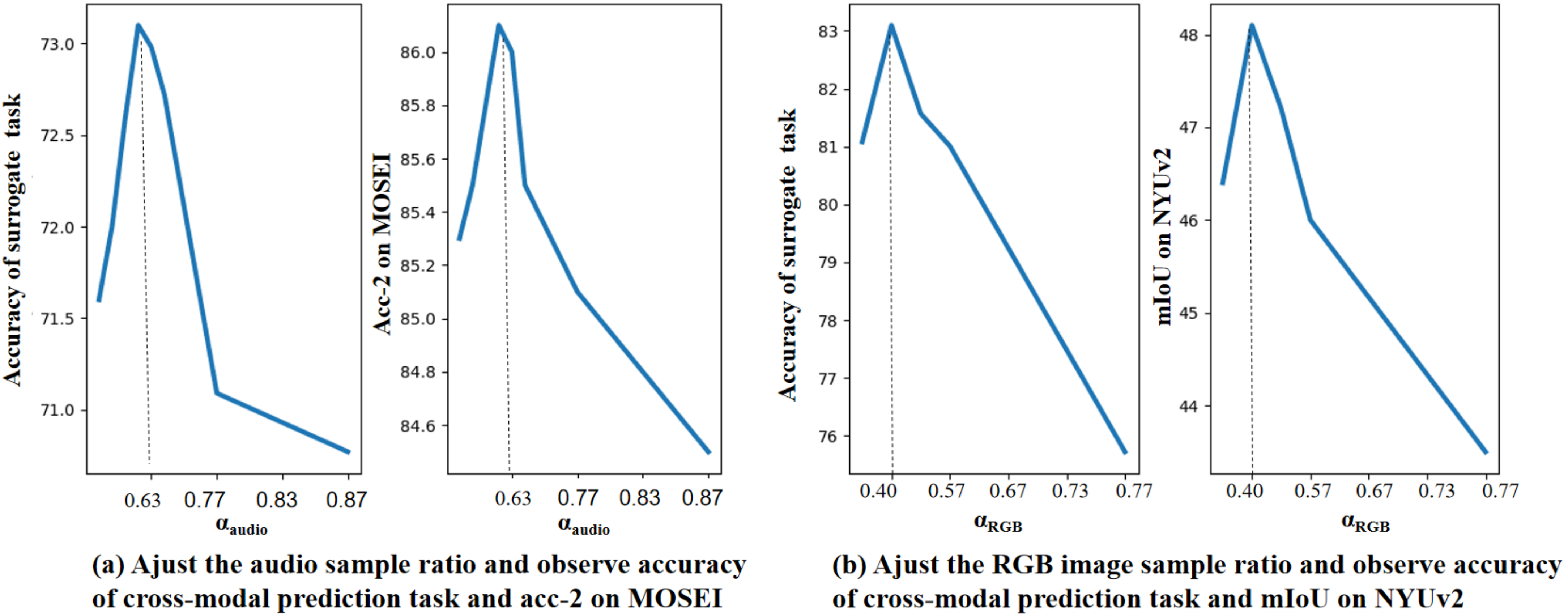}
	\caption{Correlations between the total crossmodal discrimination accuracy and the downstream task performance.}
	\label{fig:sampling}
	\vspace{-20pt}
\end{figure}

\noindent\textbf{Reward design for negative sample optimization}
We introduce crossmodal discrimination as a surrogate task for negative sample optimization in Section~\ref{sec:negopt} and argue that the total crossmodal discrimination accuracy $\mathcal{R}(\balpha)$ in Equation~\ref{eq:negreward} is a good reward function. We provide our empirical verification here. We vary the ratio $\alpha_k$ of type-$k$ negative samples while keeping the relative ratio of the rest types unchanged. We train the whole network through with the fixed negative sampling ratio and evaluate both $\mathcal{R}(\balpha)$ and the performance of the downstream task. As is shown in Figure~\ref{fig:sampling}, adjusting the proportion of different types of negative samples will influence the accuracy $\mathcal{R}(\balpha)$ of the surrogate task, which has a high correlation with downstream tasks. Too low and too high proportion for one type of negative samples both lead to low $\mathcal{R}(\balpha)$. There is a sweet spot corresponding to the best $\mathcal{R}(\balpha)$. Experiments show this sweet spot also corresponds to the best performance on downstream tasks.

\begin{figure}[t]
	\centering
	\includegraphics[width=1.0\linewidth]{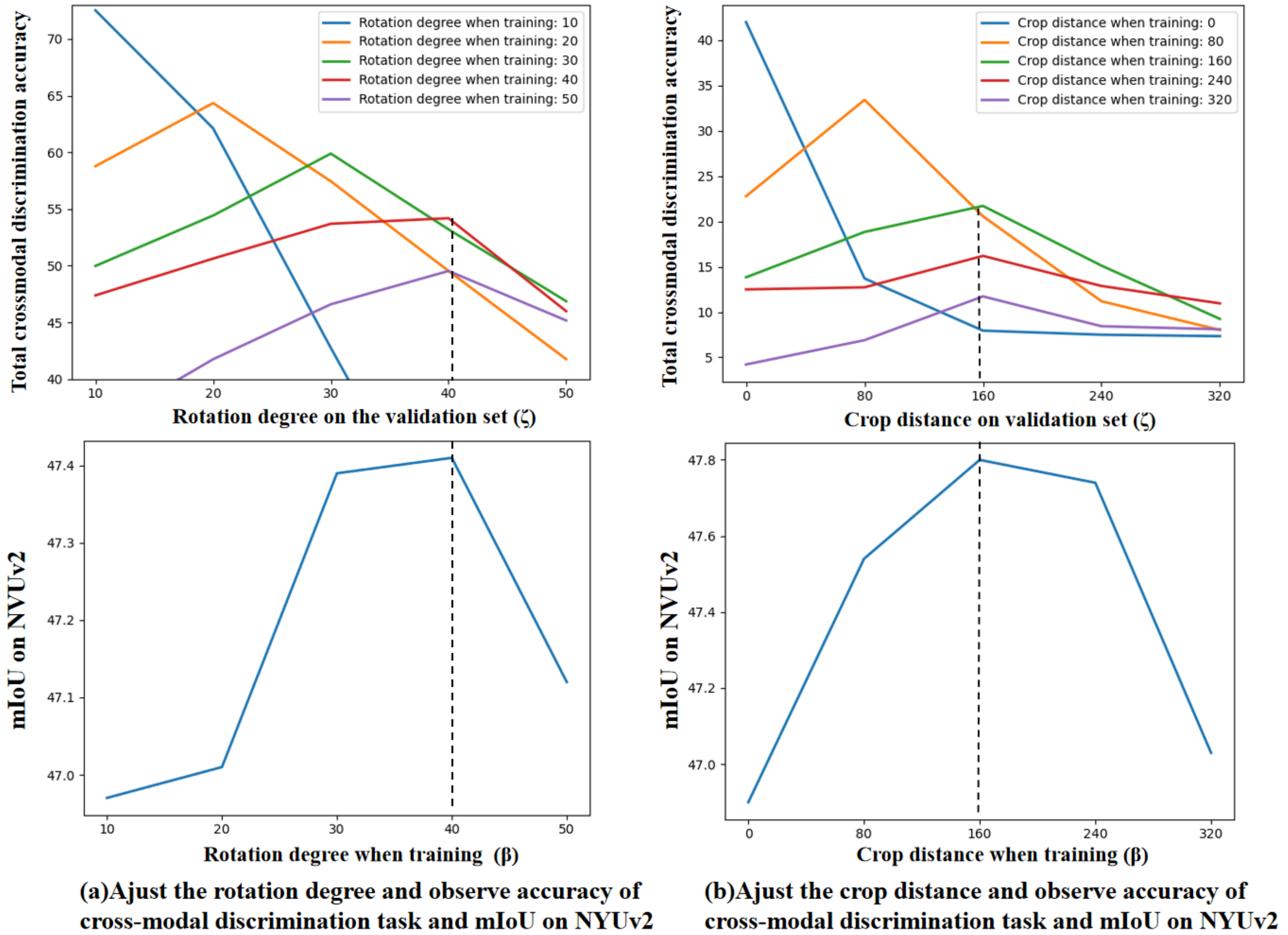}
	\caption{Empirical study justifying the reward design for postive sample optimization. In the first row we show the total crossmodal discrimination accuracy on the validation set while varying the augmentation parameter $\bzeta$ and different curves are obtained with different train time data augmentation parameters $\bbeta$. The second row shows how the performance of downstream tasks vary while changing $\bbeta$.}
	\label{fig:augmentation}
	\vspace{-20pt}
\end{figure}

\noindent\textbf{Reward design for positive sample optimization} 
Our reward function in Equation~\ref{eq:posreward} for positive sample optimization is motivated by two observations: 1). minimizing total crossmodal discrimination accuracy $\sum_{k=1}^K\mathcal{A}^k(\bg^{\balpha\bbeta},\bbeta)$ with respect to $\bbeta$ will increase the augmentation magnitude; 2). $\left\lVert\bbeta-\bzeta^*(\bbeta)\right\rVert^2$ provides cues for identifying the sweet spot beyond which larger augmentation will harm representation learning. We provide empirical studies to verify these observations in Figure~\ref{fig:augmentation}. We train networks from beginning and end with different $\bbeta$ to evaluate how the total crossmodal discrimination accuracy change while varying the data augmentation parameters $\bzeta$ on the validation set. We also evaluate how the performance of downstream tasks varies while changing the training time data augmentation parameters $\bbeta$. We experiment with two types of data augmentation - image rotation and image crop, and obtain consistent observations. $\sum_{k=1}^K\mathcal{A}^k(\bg^{\balpha\bbeta},\bbeta)$ indeed drops while increasing $\bbeta$. Moreover,  $\bzeta^*(\bbeta)$ corresponds to the peak of each curve in the first row and it is very close to $\bbeta$ when $\bbeta$ is small. Once $\bbeta$ goes beyond a sweet spot, which gives the best performance on downstream tasks, $\bzeta^*(\bbeta)$ no longer tracks the value of $\bbeta$ and $\left\lVert\bbeta-\bzeta^*(\bbeta)\right\rVert^2$ will give a penalty for further increasing $\bbeta$. In practice, we find our reward function powerful enough for identifying the best training time data augmentation parameters.


\noindent\textbf{Robustness to uninformative modality}
TupleInfoNCE emphasizes the modality which is easy to be ignored. An obvious question is whether it is robust to uninformative modalities. We conduct experiments on MOSEI multimodal sentiment analysis task and add an uninformative modality named timestamp which denotes the relative time in a sequence. Results show using these four modalities, we achieve 52.6 Acc-7, which is only $0.1\%$ lower than before. The final negative sample ratio among the four modalities is roughly 3(text): 3(video): 4(audio): 1(timestamp), showing our method successfully identifies that ``timestamp'' is not something worthy of much emphasis.

\section{Conclusion}
This paper proposes a new objective for representation learning of multimodal data using contrastive learning, TupleInfoNCE. The key idea is to contrast multimodal anchor tuples with challenging negative samples containing disturbed modalities and better positive samples obtained through an optimizable data augmentation process. 
We provide a theoretical basis for why TupleInfoNCE works, an algorithm for optimizing TupleInfoNCE with a self-supervised approach to select the contrastive samples, and results of experiments showing ablations and state-of-the-art performance on a wide range of multimodal fusion benchmarks.   

\section*{Acknowledgement}
This work was supported by the Key-Area Research and Development Program of Guangdong Province (2019B121204008) and the Center on Frontiers of Computing Studies (7100602567).

{\small
\bibliographystyle{ieee_fullname}
\bibliography{main}
}

\clearpage
\section*{\centerline{Appendix}}
\subsection*{A. Does representation quality improve as number of views increases?}
\label{sec:revisit}
To measure the quality of the learned representation, we consider the task of semantic segmentation on NYUv2. In the 1 modality setting case, TupleInfoNCE using RGB modality coincides with InfoNCE. In 2-3 modalities cases, we sequentially add depth and normal modalities.

\begin{table}[h]
	\centering
	\label{tab:revisit}
	\begin{tabular}{c|c} 
		\toprule
		Modality & mIoU\\
		\midrule 
		RGB & 40.8  \\
		RGB + depth & 44.1  \\
		RGB + normal  & 44.5 \\
		depth + normal  & 46.3  \\
		RGB + depth + normal & \textbf{48.1}\\
		\bottomrule
	\end{tabular}
	\vspace{5pt}
	\caption{We show the mean Intersection over Union (mIoU) for the NYU-Depth-V2 dataset, as TupleInfoNCE is trained with increasingly more modalities from 1 to 3. The performance steadily improves as new modalities are added.}
\end{table}

As shown in Tab.~\ref{tab:revisit}, We see that the performance steadily improves as new modalities are added. This finding is consistent with that of CMC~\cite{tian2019contrastive} who learn a cross-modal embedding space. 

\subsection*{B. Comparisons to InfoMin}
\label{sec:augmentation_search}
Infomin~\cite{tian2020makes} proposes to reduce the mutual information between different views while retaining the task-relevant information as much as possible. But how to explore task-relevant information without labels is a very challenging problem. Infomin utilizes an adversarial training strategy to search for good views in a weakly-supervised manner. When no labels are available, which is the case in a truly self-supervised representation learning setting, the effectiveness of Infomin is greatly reduced.
Taking NYUv2 semantic segmentation task as an example, in a truly self-supervised representation learning setting, the optimal rotation parameter Infomin finds is 70 degrees, which leads to 46.91 mIoU after downstream fine-turning, while the optimal rotation parameter TupleInfoNCE finds is 40 degree, corresponding to 47.41 mIoU. 


\subsection*{C. Use TupleInfoNCE in four modalities}
\label{sec:four}
We conduct another experiment on NYUv2 to see whether our method can help multi-modal semantic scene understanding in the case of 4 modalities. Following CMC, modalities we used are L, ab, depth, normal. We follow the 3 modalities setting and use random cropping, rotation, and jittering as the augmentation strategy, thus learning a fused representation by contrasting tuples that contain 4 modalities. Using 4 modalities, the best mIoU we obtain is 48.6. This is 0.5 higher than our 3 modalities baseline, showing that our method has strong generalization ability. These again validate that our TupleInfoNCE which contrasts multi-modal input tuples can produce better-fused representations.

\subsection*{D. Sampling efficiency of TupleInfoNCE}
\label{sec:ablation}
\begin{figure}[t]
	\centering
	\includegraphics[width=1.0\linewidth]{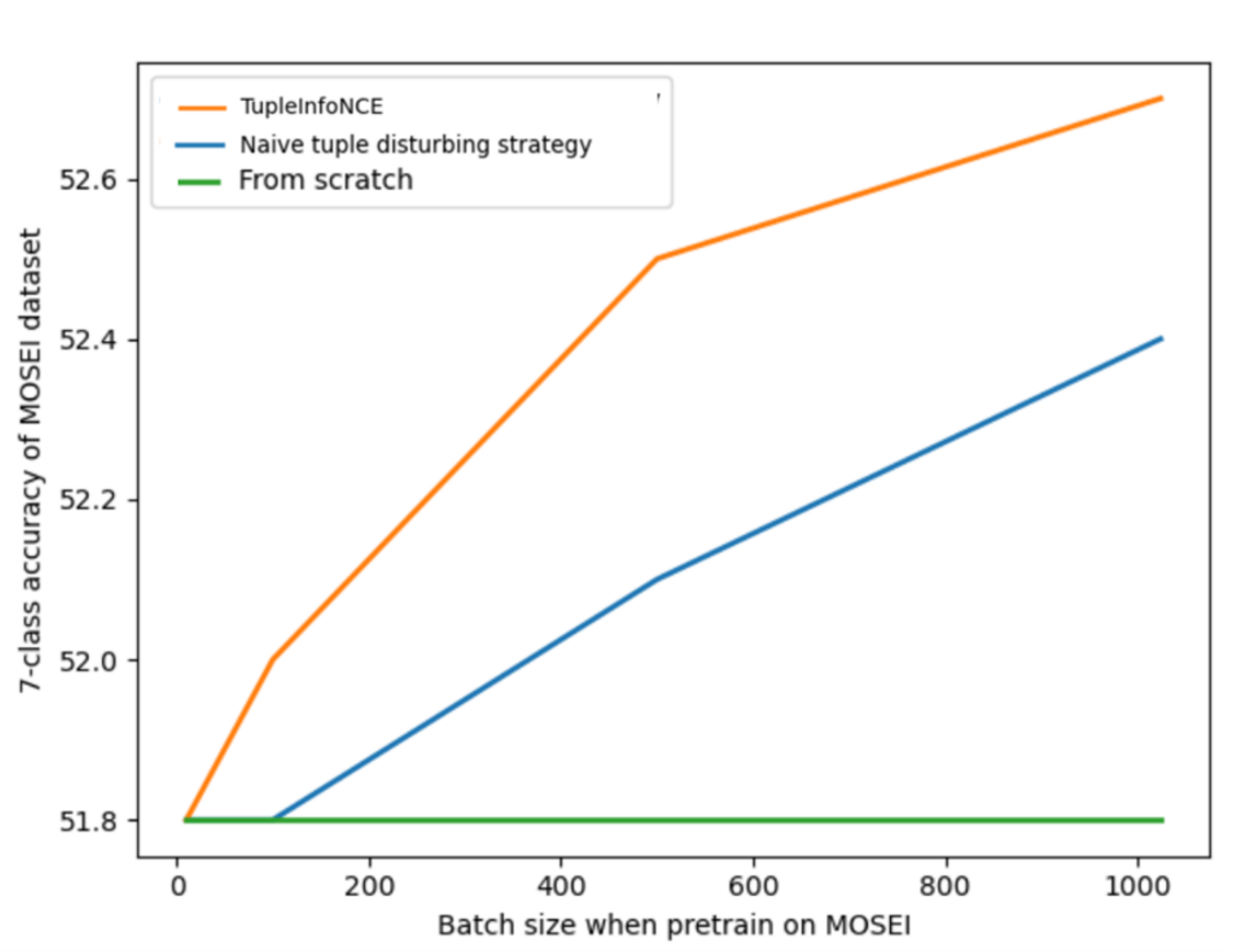}
	\caption{Sample efficiency of TupleInfoNCE}
	\label{fig:batchsize}
\end{figure}

TupleInfoNCE disturbs each modality separately to generate k disturbed negative samples. However, The most direct method is the naive sampling strategy which disturbs all modalities simultaneously to generate 1 disturbed negative sample. We conduct another experiments to compare our method with the naive sampling strategy from the perspective of efficiency. Figure~\ref{fig:batchsize} shows that our method is more efficient than naive tuple disturbing strategy. Our method with a batch size of 512 already outperforms naive tuple disturbing strategy with a batch size of 1024.

\subsection*{E. Proof for the MI lower bound of TupleInfoNCE}
\label{sec:proof}
The TupleInfoNCE loss is essentially a categorical cross-entropy classifying positive tuples $\bt_{2,i}\sim p_{\bbeta}(\bt_2|\bt_{1,i})$ from $N-1$ negatives views $\bt_{2,j}\sim q_{\balpha}(\bt_2)$. We use $p_{\balpha\bbeta}(d=i|\bt_{1,i})$ to denote the optimal probability for this loss where $[d=i]$ is an indicator showing that $\bt_{2,i}$ is the positive view. Then we can derive $p_{\balpha\bbeta}(d=i|\bt_{1,i})$ as follows:

\begin{equation*}
\begin{split}
p_{\balpha\bbeta}(d=i|\bt_{1,i}) &= \frac{p_{\bbeta}(\bt_{2,i}|\bt_{1,i})\prod_{l\neq i}q_{\balpha}(\bt_{2,l})}{\sum_{j=1}^Np_{\bbeta}(\bt_{2,j}|\bt_{1,i})\prod_{l\neq j}q_{\balpha}(\bt_{2,l})}\\
&=\frac{\frac{p_{\bbeta}(\bt_{2,i}|\bt_{1,i})}{q_{\balpha}(\bt_{2,i})}}{\sum_{j=1}^N\frac{p_{\bbeta}(\bt_{2,j}|\bt_{1,i})}{q_{\balpha}(\bt_{2,j})}}
\end{split}
\end{equation*}

\noindent It can be seen from above that the optimal value for $f(\bt_{2,j},\bt_{1,i})$ in $\mathcal{L_{\text{TNCE}}(\balpha,\bbeta)}$ is proportional to $\frac{p_{\bbeta}(\bt_{2,j}|\bt_{1,i})}{q_{\balpha}(\bt_{2,j})}$. Insert this density ratio back to $\mathcal{L_{\text{TNCE}}(\balpha,\bbeta)}$ we get:

\begin{equation*}
\begin{split}
& \mathcal{L}_{\text{TNCE}}^{\text{OPT}}(\balpha,\bbeta)\\
=& -\mathbb{E}\,\text{log}\left[\frac{\frac{p_{\bbeta}(\bt_{2,i}|\bt_{1,i})}{q_{\balpha}(\bt_{2,i})}}{\frac{p_{\bbeta}(\bt_{2,i}|\bt_{1,i})}{q_{\balpha}(\bt_{2,i})}+\sum_{j\neq i}\frac{p_{\bbeta}(\bt_{2,j}|\bt_{1,i})}{q_{\balpha}(\bt_{2,j})}}\right]\\
=&\, \mathbb{E}\,\text{log}\left[1+ \frac{q_{\balpha}(\bt_{2,i})}{p_{\bbeta}(\bt_{2,i}|\bt_{1,i})}\sum_{j\neq i}\frac{p_{\bbeta}(\bt_{2,j}|\bt_{1,i})}{q_{\balpha}(\bt_{2,j})}\right]\\
\approx&\, \mathbb{E}\,\text{log}\left[1+ \frac{q_{\balpha}(\bt_{2,i})}{p_{\bbeta}(\bt_{2,i}|\bt_{1,i})}(N-1)\underset{\bt_{2,j}}{\mathbb{E}}\frac{p_{\bbeta}(\bt_{2,j}|\bt_{1,i})}{q_{\balpha}(\bt_{2,j})}\right]\\
=&\, \mathbb{E}\,\text{log}\left[1+ \frac{q_{\balpha}(\bt_{2,i})}{p_{\bbeta}(\bt_{2,i}|\bt_{1,i})}(N-1)\right]\\
\geq& \, \mathbb{E}\,\text{log}\left[\frac{q_{\balpha}(\bt_{2,i})}{p_{\bbeta}(\bt_{2,i}|\bt_{1,i})}N\right]\\
\end{split}
\end{equation*}

\noindent We design our negative ``proposal'' distribution as $q_{\balpha}(\bt_2)=\alpha_0p(\bt_2)+\sum_{k=1}^K\alpha_kp(\bar{\bv}_2^k)p(\bv_2^k)$. Insert this to the inequality above we obtain:

\begin{equation*}
\begin{split}
& \mathcal{L}_{\text{TNCE}}^{\text{OPT}}(\balpha,\bbeta)\\
\geq& \, \mathbb{E}\,\text{log}\left[\frac{\alpha_0p(\bt_{2,i})+\sum_{k=1}^K\alpha_kp(\bar{\bv}_{2,i}^k)p(\bv_{2,i}^k)}{p_{\bbeta}(\bt_{2,i}|\bt_{1,i})}N\right]\\
\geq& \, \mathbb{E}\,\text{log}\left[\frac{(p(\bt_{2,i}))^{\alpha_0}\prod_{k=1}^K(p(\bar{\bv}_{2,i}^k)p(\bv_{2,i}^k))^{\alpha_k}}{p_{\bbeta}(\bt_{2,i}|\bt_{1,i})}N\right]\\
=& \, \mathbb{E}\,\text{log}\left[\frac{p(\bt_{2,i})}{p_{\bbeta}(\bt_{2,i}|\bt_{1,i})}\prod_{k=1}^K(\frac{p(\bar{\bv}_{2,i}^k)p(\bv_{2,i}^k)}{p(\bt_{2,i})})^{\alpha_k}N\right]\\
\approx & \, \mathbb{E}\,\text{log}\left[\frac{p_{\bbeta}(\bt_{2,i})}{p_{\bbeta}(\bt_{2,i}|\bt_{1,i})}\prod_{k=1}^K(\frac{p(\bar{\bv}_{2,i}^k)p(\bv_{2,i}^k)}{p(\bt_{2,i})})^{\alpha_k}N\right]\\
=& \, \text{log}(N)-I(\bt_{2,i};\bt_{1,i}|\bbeta)-\sum_{k=1}^K\alpha_kI(\bv_{2,i}^k;\bar{\bv}_{2,i}^k)
\end{split}
\end{equation*}

\noindent Therefore $I(\bt_{2,i};\bt_{1,i}|\bbeta)+\sum_{k=1}^K\alpha_kI(\bv_{2,i}^k;\bar{\bv}_{2,i}^k)\geq \text{log}(N)-\mathcal{L}_{\text{TNCE}}^{\text{OPT}}(\balpha,\bbeta)$.

\subsection*{F. Examples for negative sample optimization}
\label{sec:sampling}
This paper sets out with the aim of assessing the importance of complementary synergies between modalities. What is surprising is that increasing the ratio of weak modalities which is considered to contain less useful information can produce better-fused multi-modal representations. To be specific, the negative sample sampling ratio is roughly 1(RGB): 2(depth): 3(normal) in NYUv2 semantic segmentation task, 1(text): 1(video): 4(audio) in MOSI/MOSEI sentiment analysis task, and 1(point cloud): 3(RGB color): 2(height) in SUNRGB-D 3D Object Detection task. These results further support our idea of contrasting multi-modal input tuples to avoid weak modalities being ignored and to facilitate modality fusion. A possible explanation for this might be that our proposed TupleInfoNCE encourages neural networks to use complement information in weak modalities, which can avoid networks converging to a lazy suboptimum where the network relies only on the strongest modality. This observation may support the hypothesis that mining useful information contained in weak modalities instead of always emphasizing strong modal information can obtain better fusion representations.

\subsection*{G. Examples for positive sample optimization}
\label{sec:augmentation}
As mentioned in the main paper, we need to optimize the positive samples so that the contrastive learning can keep the shared information between positive and anchor samples while abstracting away nuisance factors. In the NYUv2 semantic segmentation task, rotation of 40 degrees, cropping with 160 center pixels, and Gaussian noise with a variance of 50 is best for RGB modality, while no augmentation is better for depth and normal modalities. As for the SUNRGB-D 3D Object Detection task, the final augmentation is to rotate the point cloud by 10 degrees, add Gaussian noise with a variance of 30 to the RGB image, and apply Gaussian noise with a variance of 10 to the height modality.

\subsection*{H. Dropout training}
\label{sec:dropout}
\vspace*{2mm}\noindent\textbf{Dropout training}
Note $\bg^{\balpha\bbeta}$ is originally designed to consume all input modalities, while in the crossmodal discrimination task, $\bg^{\balpha\bbeta}$ needs to handle missing modalities as shown in Equation. We adopt a simple \textit{dropout training} strategy to achieve this goal. To be specific, we randomly mask out modalities and fill them with placeholder values in the input. The missing modalities are the same in the positive and negative samples, yet could be different in the anchor tuple. This dropout strategy is only adopted with a probability of 0.6 for each training batch and the rest of the time we feed complete inputs to the feature encoder.

\paragraph{Robustness to dropout training}
We found dropout training does not cause a drop in feature quality compared to non-dropout training. 
We first obtain the optimal hyper-parameter by grid searching. When we use a dropout training strategy with fixed optimal hyper-parameter in NYUv2 semantic segmentation task, we achieve 48.3 mIoU performance which only outperforms our strategy by $0.2\%$, which presumably means dropout training might distill information from other modalities to avoid representation quality degradation. 

\subsection*{I. Implementation Details}
\label{sec:implementation_details}
\begin{figure}[t]
	\centering
	\includegraphics[width=\linewidth]{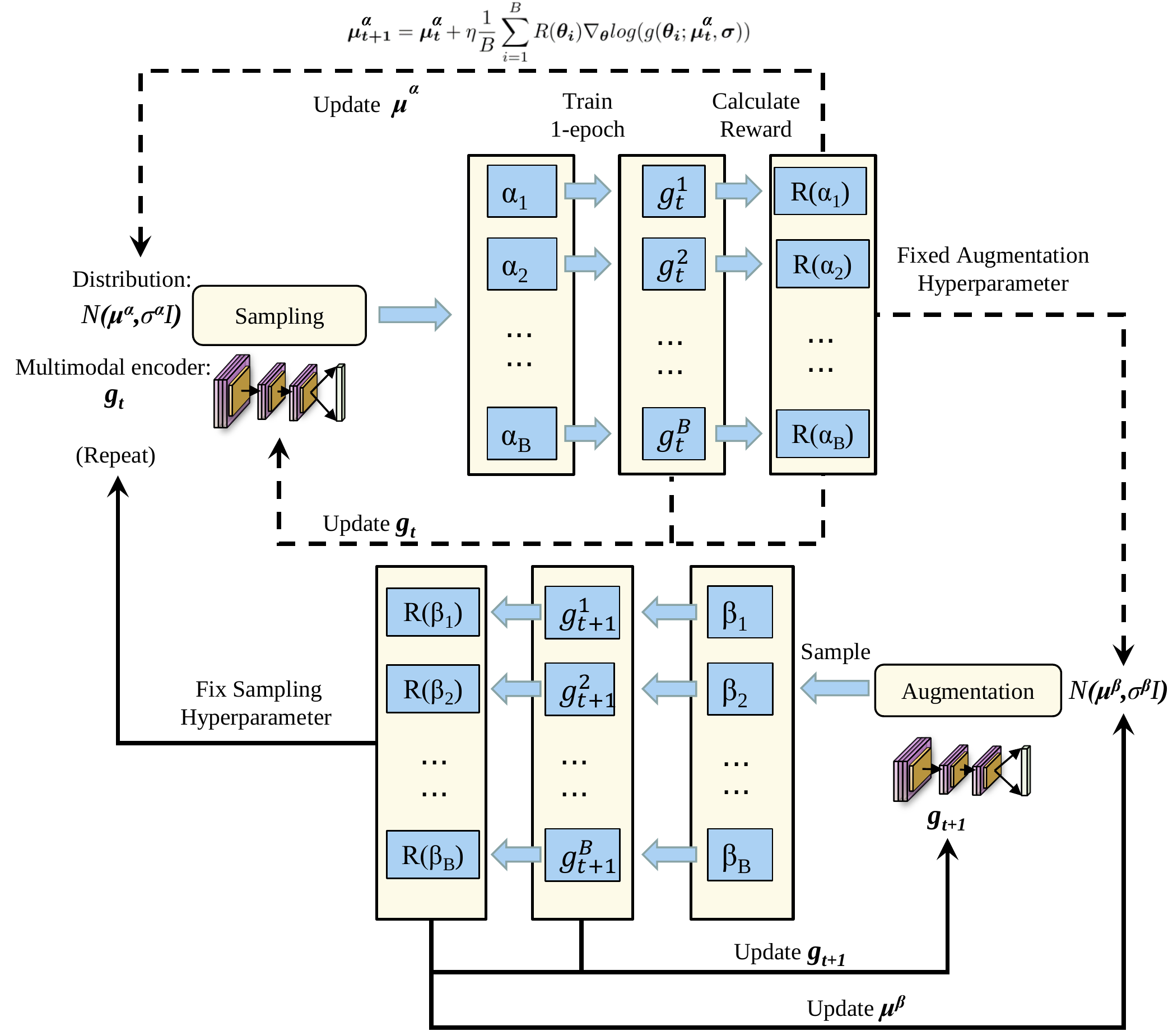}
	\caption{Overview of sample optimization.}
	\label{fig:optimaize}
	\vspace{-15pt}
\end{figure}

For NYU-Depth-V2 semantic segmentation task, we train the model with one V100 GPU for 500 epochs. The batch size is 20. We use SGD+momentum optimizer with an initial learning rate 0.25. We use onecycle learning rate scheduler. We set the weight decay as 1e-4. In the experiment of SUN RGB-D 3D object detection, our settings are consistent with VoteNet. We train the model on one 2080Ti GPU for 180 epochs. The initial learning rate is 0.001. We sample 20,000 points from each scene and the voxel size is 5cm. As for sentiment analysis on MOSEI and MOSI,  We used the same data as MISA, where text features are extracted from BERT. Batch size is 32 for MOSI, and 16 for MOSEI.
For sample optimization details, we provide a flow chart Figure~\ref{fig:optimaize} to further support the main paper.

\typeout{get arXiv to do 4 passes: Label(s) may have changed. Rerun}
\end{document}